\documentclass[11pt,a4paper]{llncs}
\usepackage{amsmath}
\usepackage{amssymb}
\usepackage{graphicx}
\usepackage{marvosym}
\usepackage{url}
\usepackage{hyperref}

\usepackage{listings}
\usepackage{color}

\definecolor{dkgreen}{rgb}{0,0.6,0}
\definecolor{gray}{rgb}{0.5,0.5,0.5}
\definecolor{mauve}{rgb}{0.58,0,0.82}

\usepackage{fancyhdr}

\usepackage{geometry}
\newcommand{\keywords}[1]{\par\addvspace\baselineskip
\noindent\keywordname\enspace\ignorespaces#1}

\fancypagestyle{firstpage}{\fancyhf{}
}

\begin{document}


\title{\LARGE{End-to-End Bangla AI for Solving Math Olympiad Problem Benchmark:Leveraging Large Language Model Using Integrated Approach}}


%
%
\author{\large{H.M.Shadman Tabib  \and Jaber Ahmed Deedar }}
\institute{\large{Department of Computer Science and Engineering,Bangladesh University of Engineering and Technology\\ Dhaka,Bangladesh}}

%


%
%


\maketitle

\thispagestyle{firstpage}

\begin{abstract}
This work introduces systematic approach for enhancing large language models (LLMs) to address Bangla AI mathematical challenges. Through the assessment of diverse LLM configurations, fine-tuning with specific datasets, and the implementation of Retrieval-Augmented Generation (RAG), we enhanced the model's reasoning precision in a multilingual setting. Crucial discoveries indicate that customized prompting, dataset augmentation, and iterative reasoning improve the model's efficiency regarding Olympiad-level mathematical challenges.
\end{abstract}


\section{Introduction}

In recent years, large language models (LLMs) have exhibited tremendous potential in tackling hard mathematics problems, including ones at Olympiad-level complexity. Omni-math is a global benchmark that covers several sub-domains and difficulty levels\cite{b1}. Odyssey math data measures mathematical problem-solving ability in LLMs\cite{b2}. These are just a few of the tools that researchers have made to test and improve model reasoning. Researchers have devised numerous methodologies and standards to evaluate and enhance the effectiveness of these models in mathematical reasoning.

A significant method is the self-critique pipeline, implemented in the ChatGLM-Math model. This approach entails training a Math-Critique model based on the LLM to provide feedback signals. The LLM's own outputs then undergo rejective fine-tuning and direct preference optimization for data acquisition. Experiments with the ChatGLM3-32B model showed that this pipeline greatly enhances the ability to solve mathematical problems while maintaining language skills, making it better than larger LLMs. \cite{b5}

Another new technique involves the creation of MetaMath, a refined language model focused on mathematical reasoning. The MetaMath methodology initiates by formulating mathematical inquiries from several viewpoints, culminating in a novel dataset termed MetaMathQA. This dataset used to fine-tune the LLaMA-2 models. Studies using benchmarks like GSM8K and MATH show that MetaMath performs better than a number of open-source LLMs. For example, the MetaMath-7B model achieved a GSM8K precision of 66.5\% and a MATH accuracy of 19.8\%, which is much better than the performance of models of the same size. \cite{b6}

New standards have been established to assess the mathematical reasoning capabilities of LLMs, in addition to model-specific improvements. Mathador-LM is a dynamic benchmark derived from the Mathador game, wherein the goal is to get a target number through fundamental arithmetic operations applied to a specified set of base numbers, adhering to straightforward criteria. This test integrates ruleset interpretation, planning, and problem-solving, offering a thorough evaluation of LLMs' mathematical reasoning ability. Research indicates that modern models do poorly on Mathador-LM, achieving scores much below those of average third graders, underscoring the necessity for enhancements in this domain.\cite{b7}

Additionally, ConceptMath is a multilingual benchmark (English and Chinese) that evaluates concept-specific mathematical reasoning in LLMs. ConceptMath organizes math problems in a way that lets them be tested at different levels of detail with concept-specific accuracy. This is different from traditional benchmarks that only measure general mathematical reasoning with average accuracy. Evaluations have shown that existing LLMs, while achieving high average accuracies on common metrics, exhibit significant performance variations among different mathematical ideas and may even struggle with basic ones.\cite{b8}

Another paper \cite{b9} introduces the MCT Self-Refine (MCTSr) algorithm, a novel approach integrating Large Language Models (LLMs) with Monte Carlo Tree Search (MCTS) to improve performance in complex mathematical reasoning tasks. The MCTSr algorithm addresses the challenges of accuracy and reliability in LLM-based reasoning by employing systematic exploration and heuristic self-refinement mechanisms. Another work presents OlympiadBench, a bilingual multimodal benchmark consisting of 8,476 Olympiad-level mathematics and physics tasks, intended to assess the advanced reasoning abilities of Large Language Models (LLMs) and Large Multimodal Models (LMMs). Preliminary assessments indicate considerable difficulties, with the top-performing model, GPT-4V, attaining merely 17.97\% accuracy, highlighting the benchmark's stringency and its potential as a tool for furthering AGI research.\cite{b10}

These advancements highlight the continuous endeavors to improve the mathematical problem-solving capabilities of LLMs via novel training methodologies and thorough assessment standards.Still, there are difficulties making sure LLMs can operate across several languages and problem types—especially for challenging tasks in non-English environments. Emphasizing the continuous improvement of these models, efforts including multilingual machine translation\cite{b3} and optimized frameworks like BEATS, which strengthens mathematical reasoning through back-verification and adaptive disambiguation\cite{b4}, show this study builds on previous research by concentrating on fine-tuning LLMs to solve Bangla math problems. It does this by using a structured strategy of dataset enhancement, retrieval-augmented generation (RAG) \cite{rag}, and personalized prompting to get the best results for Bangla AI math problems.
\vspace{0.2cm} \\

\section{Methodology}

\subsection{Model Selection}
In our AI-driven Bangla Math Olympiad solution, we tested many large language models (LLMs) to find the best one for mathematical reasoning in Bangla. Our selection process evaluated models based on their mathematical problem-solving, Bangla flexibility, and computing economy.

Specifically designed for mathematical problem-solving, \textbf{Deepseek-Math-7b-Instruct} \cite{deepseek}was evaluated. It did not perform well  in our used dataset.

Next, we evaluated the \textbf{NuminaMath} \cite{numinamath} models, which include \textbf{NuminaMath-7B-TIR} and \textbf{NuminaMath-7B-CoT} .These models were used to solve the problems in AI Mathematical Olympiad organized by AIMO. This model performs moderately well on our used dataset, but not up to expected level due to language difference.

We then focused on the \textbf{Qwen2.5-7B-Instruct, Coder, and Math Instruct} models\cite{qwen}. Our Bangla evaluations of the Qwen2.5-7B-Math Instruct model for mathematical exercises were not enough satisfactory. Qwen2.5-7B-coder, which is transcribed for coding, also failed to generate codes for most Bangla languages. Manual inference for some selected problems revealed that they could not solve trivial problems. 

However, Qwen2.5-7B-Instruct understood and solved Bangla maths problems well. The model's versatility and Bangla language adaptability made it excellent for our purpose. It balanced mathematical thinking with Bangla language adaptation better for our purposes after extensive evaluations. Next, we tested on Qwen2.5-32B-Instruct-AWQ without fine-tuning or RAG, which makes a more improved accuracy over test set marking 77 out of 100 showing best of all. It implies that using model trained with large parameter ensures better performance.

\subsection{Datasets}

\begin{itemize}

    \item \textbf{BDMO Dataset} \\
    This dataset is provided by Bangladesh Math Olympiad (BDMO). This dataset originally contains 209 Bangla Math Olympiad problems with their numerical solutions. We extended the dataset by adding step-by-step TIR solutions. We also translated the problem statements into English for further testing. There are also 100 test problems available in this kaggle contest \href{https://www.kaggle.com/competitions/dlsprint3}{\textbf{here}} upon which the accuracy score is calculated.

    \item \textbf{Translated Numina CoT Dataset:} \\
    This dataset comprises 0.8M problems formatted using the Chain of Thought reasoning technique, focusing on step-by-step solutions which were used to train NuminaMath-7B-CoT in AI-MO Competition. For our use case, these were translated using gemini-1.5-flash model into Bangla for fine-tuning. It supports the training of models capable of multi-step reasoning. This contains both problem statement and problem solution step description in Bangla. \\
    
    \item \textbf{Translated Numina TIR Dataset:} \\
    A foundational dataset with around 72K math competition problems ,with each solution generated by GPT-4 with tool integrated reasoning (TIR)\cite{tir}. Next, these were also translated likewise for our convenience. \\
    
    \item \textbf{Synthetic Problems Dataset:} \\
    This dataset includes a variety of synthetic mathematical problems generated using the GPT4o API based on BDMO dataset. These problems ensure multiple similar problems with paraphrasing in order to  ensure diversity while training for each category. \\
\end{itemize}
These datasets can be found in the \hyperref[dataset_availability]{Dataset Availability} section.

\subsection{Preprocessing}
We underwent a comprehensive data preprocessing method that integrated Retrieval-Augmented Generation (RAG) with keyword search to solve the  challenges associated with Bangla Math Olympiad. We initially identified significant keywords from each mathematical issue manually, focusing on terms that focus on the essential concepts and categories of the problems. Utilizing these keywords, we conducted similarity searches inside our  problem dataset to identify issues with analogous structures or themes. The recognized similar problems and their solutions as Tool Integrated Reasoning were then incorporated as few-shot instances in the model's prompts, providing contextual assistance and enhancing the model's problem-solving capabilities. By integrating RAG with keyword-based retrieval, we enhanced the model's ability to comprehend and address Bangla Math Olympiad problems effectively. This approach associated contextual examples with each problem, assists in model's effectiveness in solving Math problems.

\subsection{Augmentation}\label{AA}
We expanded  and diversified our training dataset in order to enhance the capabilities of our model with the utilization of data augmentation techniques, to solve the challenges presented by the Bangla Math Olympiad. With the help of OpenAI's GPT-4.0, we generated five paraphrased and similar versions for each of the original problems. These ensure that the original difficulties' complexities and nuances were maintained. After the implementation of this modification, heterogeneity was introduced into the dataset, which involve manual categorization over the problem sets. By exposing the model to a greater variety of problem statements, we were able to improve its flexibility and robustness through the process of addressing a variety of mathematical challenges in Bangla .

\subsection{Fine-tuning}
Fine-tuning of the Qwen-2.5-7B-Instruct model was done using \textbf{1x NVIDIA H100 NVL GPU} for efficient training. The model underwent three phases to improve its reasoning for Bangla AI mathematical problems and solutions. The first phase involved fine-tuning the model using the translated Numina TIR dataset, which had around 72,000 problems and step-by-step python solutions. The second stage involved using the Numina CoT dataset, which included over around 800,000 CoT problem solution pairs. For our use case, we used 250,000 of those. Then, we used GPT4o API for creating synthetic datasets to increase the model's adaptability and versatility. The final stage focused on paraphrasing and augmentation, rephrasing problem sets to simulate different mathematical presentations. The model was evaluated using BDMO dataset and test dataset at each level to examine accuracy improvements and fine-tuning stage effects. Several rounds of model training improved the model's ability to understand complex reasoning processes and improve its Bangla AI mathematics comprehension and response formulation. All the models fine-tuned in each step can be found in the \hyperref[model_availability]{Model Availability} section.
\begin{figure}[h!]
\centering
\includegraphics[width=0.8\textwidth]{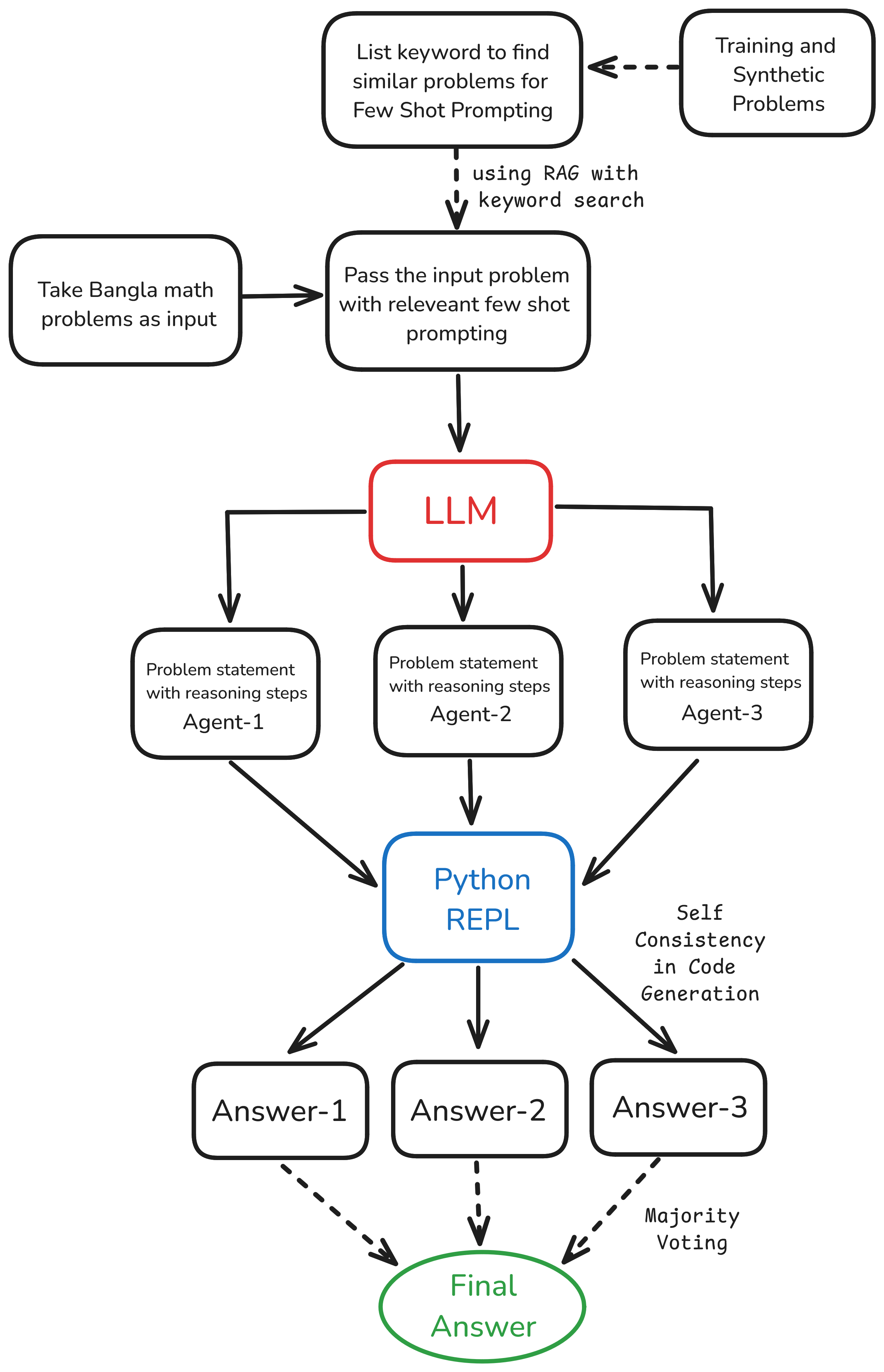} 
\caption{Model Architecture}
\label{fig:model_architecture}
\end{figure}

\subsection{Model Architecture and Flow}
Figure ~\ref{fig:model_architecture} illustrates the architecture of our system developed to analyze the Bangla mathematical problem. The system employs a Large Language Model (LLM) in conjunction with TIR agents and self-consistency \cite{self}. The TIR agents generate python code iteratively until the correct solution is found and then executes them in REPL (Read Eval Print Loop). According to self-consistency prompting, final answer is chosen by majority voting among the differently generated solutions.

The procedure starts by accepting Bangla mathematics questions as input. Keywords are derived analyzing the training data  and a Retrieval-Augmented Generation (RAG) technique is employed to identify analogous instances.They are basically categorized as similar problems matching their keywords and then used these similar problems for few shot prompts. These examples assist the LLM in producing methodical solutions. Multiple TIR agents develop distinct solution paths to investigate several methods for addressing the challenge, subsequently testing them in a Python code execution environment.

Upon the failure of a solution path, the system re-evaluates the problem to rectify any inaccuracies. Ultimately, the system aggregates the agents' results, employing a voting mechanism to choose the optimal option. This methodology guarantees precision and dependability through the integration of reasoning, verification, and consensus techniques.

\subsection{Model Inference and Prompting}
The TIR agents were tested using system prompts containing no instruction and  prompts containing tailored instructions. These instructions were based on the problems' categories which were manually curated investigating the training dataset. Such of these can be found \hyperref[prompting]{here}.  Moreover, we tested both zero shot and few shot prompting techniques. For few shot examples, we used keyword search RAG where problems are categorized matching relevant keywords.

\subsection*{Prompts Used}
\subsection*{Base Prompt}
\begin{lstlisting}
Here is a math problem in Bengali:
{problem}

The answer is a non-negative integer. Please reason step by step to solve the problem above. Provide Python code to verify your reasoning.

Put your final integer answer within \boxed{}.
\end{lstlisting}

\subsection*{Tailored Instruction for Number Theory Problems}
\begin{lstlisting}
Please write brute force solution codes to answer.
\end{lstlisting}

\subsection*{Translation Prompt}
\begin{lstlisting}
Please translate the problem to English first. Then solve the problem.
\end{lstlisting}

\subsection*{Advanced Prompt}
\begin{lstlisting}
Solve the problem in Bengali by explaining all intermediate steps in English. 
Use Python to verify the solution. 
Highlight important mathematical reasoning.
\end{lstlisting}

\subsection*{Step-by-Step Guidance Prompt}
\begin{lstlisting}
Explain the problem-solving process using detailed reasoning.
Start from basics and gradually solve the problem step by step. Show intermediate results 
clearly. 
Verify with Python.
\end{lstlisting}

\section{Results}

\begin{table}[h!]
\centering
\scriptsize 

\renewcommand{\arraystretch}{1.5} 
\setlength{\tabcolsep}{12pt} 
\begin{tabular}{|p{3.5cm}|c|p{1cm}|p{1.5cm}|c|}
\hline
\textbf{Model Configuration} & \textbf{Temperature} & \textbf{Problem Language} & \textbf{Reasoning Language} & \textbf{Result} \\
\hline
GPT4o - basic & 0.7 & Bangla & English & 115 / 209 \\
\hline
GPT4o - basic & 0 & Bangla & English & \textbf{122*} / 209 \\
\hline
GPT4o - basic & 0.7 & Bangla & Bangla & 109 / 209 \\
\hline
GPT4o - basic & 0 & Bangla & Bangla & 107 / 209 \\
\hline
GPT4o - TIR & 0 & Bangla & English & 125 / 209 \\
\hline
GPT4o - TIR & 0 & English & English & \textbf{130**} / 209 \\
\hline

\end{tabular}
\vspace{5 pt}
\caption{Comparison of GPT-4o Configurations for BDMO dataset problems. Here "basic" means direct prompting without using any tool and "TIR" denotes model is coupled with tool integrated reasoning.}
\label{table:gpt4o_comparison1}
\end{table}

\begin{table}[h!]
\centering
\scriptsize 

\renewcommand{\arraystretch}{1.5} 
\setlength{\tabcolsep}{12pt} 
\begin{tabular}{|p{3.5cm}|c|c|c|}
\hline
\textbf{Model Configuration} & \textbf{Problem Language} & \textbf{Reasoning Language} & \textbf{Result} \\
\hline
GPT4o - Basic & Bangla & Bangla & 50 / 100 \\
\hline
GPT4o - Basic & English & Bangla & 62 / 100 \\
\hline
GPT4o - Basic & Bangla & English & \textbf{64*} / 100 \\
\hline
GPT4o - Basic & English & English & 60 / 100 \\
\hline
GPT4o - TIR & Bangla & Bangla & 63 / 100 \\
\hline
GPT4o - TIR & English & Bangla & 61 / 100 \\
\hline
GPT4o - TIR  & Bangla & English & 67 / 100 \\
\hline
GPT4o - TIR  & English & English & \textbf{71**} / 100 \\
\hline

\end{tabular}
\vspace{5 pt}
\caption{Comparison of GPT-4o Configurations for test problems. Here "basic" means direct prompting without using any tool and "TIR" denotes model is coupled with tool integrated reasoning.}
\label{table:gpt4o_comparison2}
\end{table}

\begin{table}[h!]
\centering
\scriptsize 

\renewcommand{\arraystretch}{1.5} 
\setlength{\tabcolsep}{12pt} 
\begin{tabular}{|p{3.5cm}|c|c|c|}
\hline
\textbf{Model Configuration} & \textbf{Problem Language} & \textbf{Reasoning Language} & \textbf{Result} \\
\hline
Deepseek-Math-7B-Instruct-Basic & Bangla & Bangla & 28 / 100 \\

\hline

\end{tabular}
\vspace{5 pt}
\caption{Comparison of DeepSeekMath model for test problems. Here "basic" means direct prompting without using any tool.}
\label{table:gpt4o_comparison}
\end{table}
\begin{table}[h!]
\centering
\scriptsize 
\renewcommand{\arraystretch}{1.8} 
\setlength{\tabcolsep}{15pt} 
\begin{tabular}{|p{1.5cm}|c|p{3cm}|p{3cm}|}
\hline
\textbf{Count of TIR Agents} & \textbf{Instruction Given} & \textbf{Language} & \textbf{Result} \\
\hline
42 & NO        & Bengali           & 59 / 100 \\
\hline
10 & NO       & English           & 62 / 100\\
\hline
13 & NO        & English           & 53 / 100\\
\hline
10 & YES      & English           & \textbf{64*} / 100\\
\hline
5 & YES       & English           & 62 / 100\\
\hline
5 & YES       & Bengali           & 50 / 100\\
\hline
\end{tabular}
\vspace{10pt}
\caption{Benchmark Results for Numina-7B-TIR Model. Here, "no instructions" implies no system prompt was used during inference, while "instructions given" denotes tailored system prompts.}
\label{table:numina_7b_tir}
\end{table}

\begin{table}[h!]
\centering
\scriptsize 
\renewcommand{\arraystretch}{1.8} 
\begin{tabular}{|p{9cm}|p{1.5cm}|c|c|c|c|}
\hline
\textbf{Model Configuration} & \textbf{Problem Language} & \textbf{Samples} & \textbf{Depth} & \textbf{Score} \\
\hline
Qwen2.5-7B-Math-Instruct & English  & 10 & 10 & 59 \\
\hline
Qwen2.5-7B-Math-Instruct & English  & 5 & 10 & 57 \\
\hline
Qwen2.5-7B-Coder Instruct & English  & 5 & - & 65 \\
\hline
Qwen2.5-7B-Instruct & English & 5 & 5 & 70 \\
\hline
Qwen2.5-7B-Instruct & English  & 50 & 9 & 70 \\
\hline
Qwen2.5-7B-Instruct & Bangla & 50 & 9 & 68 \\
\hline
Qwen2.5-7B-Instruct (With prompt to translate in English first) & Bangla & 50 & 9 & 64 \\
\hline
Qwen2.5-32B-Instruct-AWQ & Bangla & 10 & 4 & \textbf{77*}\\
\hline
\end{tabular}
\vspace{5 pt}
\caption{Performance Comparison of Qwen2.5 Family Models for Solving Bangla AI Math Problems}
\end{table}
\begin{table}[h!]
\centering
\scriptsize 
\renewcommand{\arraystretch}{1.8} 
\setlength{\tabcolsep}{12pt} 
\begin{tabular}{|p{4cm}|c|c|c|c|}
\hline
\textbf{Dataset} & \textbf{Data Size} & \textbf{Epochs} & \textbf{Augmentation} & \textbf{Score} \\
\hline
TIR       & 72K                 & 1               & No                    & 70            \\
\hline
CoT               & 0.25M               & 1               & No                    & 40            \\
\hline
TIR                            & 72K TIR       & 3               & No                    & 70            \\
\hline
CoT + TIR                      & 72K TIR+ 0.25M CoT & 5           & Yes                   & 68            \\
\hline
CoT + TIR + RAG                & 72K TIR+ 0.25M CoT  & 5           & Yes                   & \textbf{71*}            \\
\hline
TIR + RAG                      & 72K TIR         & 5               & Yes                   & \textbf{71*}              \\
\hline
\end{tabular}
\vspace{10pt}
\caption{Performance Comparison of Fine-Tuned Models of Qwen2.5-7B-Instruct with Different Datasets for Solving Bangla AI Math Problems}
\label{table:model_comparison}
\end{table}

The study aimed to improve Bangla AI mathematics performance by finding the optimum model configuration, tweaking it, and applying RAG. The initial stage involved testing GPT-4o for reasoning and linguistic proficiency. GPT4o-TIR had the highest accuracy of 130 out of 209 in BDMO dataset. When Bangla was the main language, the model performed worse, suggesting the need for more advanced and versatile model. The second stage examined how number of TIR agents and instructional setups affected the Numina-7B-TIR model's performance. The maximum score was 64 out of 100 (here we used the test dataset) with 10 agents per problem and clear English direction as prompt. Next, we tested Qwen2.5-7B-Instruct. It was more versatile and effective arrangement, earning 70 with 5 agents and 5 depth. In the final testing phase, data augmentation with TIR, CoT, and RAG datasets enhanced model accuracy a little bit,increased up to 71. Next at the end, a larger variant of Qwen, named Qwen2.5-32B-Instruct-AWQ outperformed all other remaining models in solving problems signifying the fact that it improves with parameter enhancement. This one might be improved too with help of our additional fine-tuning,RAG and prompt techniques.The study concluded that retrieval-augmented methods and well-curated datasets improve model performance in multilingual, specialized domains.
\section{Experiments and Discussion}
In this section, we discuss the various experiments conducted to evaluate the model’s performance in solving Bangla AI math problems. Each experiment focuses on specific aspects of problem-solving, including problem categorization, prompt structuring, multilingual reasoning, and response nuances. These insights provide valuable guidelines for optimizing the model’s performance on different types of mathematical problems in Bangla. From \textbf{Table~\ref{table:gpt4o_comparison1}} and \textbf{Table~\ref{table:gpt4o_comparison2}},it is further observed that if the problem statement is in Bangla and the solution steps are described in English, there is an enhanced accuracy score over the datasets while using direct prompting without any tool integration.

\subsection{Problem Categorization for Improved Performance}
We observed that categorizing problems into specific types improves the model's ability to generate accurate responses. When problems are grouped by category—such as Number Theory, Geometry, and Combinatorics—the model benefits from prompts tailored to each category’s unique solving techniques.

\subsection{Tailored Prompting for Problem Types}
\label{prompting}
Specific types of problems showed better performance when prompted with targeted solving techniques:
\begin{itemize}
    \item \textbf{Number Theory}: The model performs better on number theory problems if prompted to solve using brute force methods.
    \item \textbf{Geometry}: In some cases, geometry problems are more effectively solved when the prompt includes instructions to approach the solution with coordinate geometry transformations.
    \item \textbf{Combinatorics and Functional Equations}: Problems involving combinatorics, functional equations, and recursion show improved performance when prompted with hints about dynamic programming techniques.
\end{itemize}

\subsection{Multilingual Reasoning and Repetitive Querying}
In general, the model achieves higher accuracy in English than in Bangla. However, during experimentation, we found that if multiple queries are submitted—one translated into English and the other kept in its original Bangla form—the model can enhance its reasoning capabilities through repetitive reasoning. This approach allows the model to analyze the problem in different linguistic contexts, resulting in more accurate solutions.

\subsection{Prompt Phrasing and Politeness}
Interestingly, a few cases revealed that the model follows prompts more strictly if the word “please” is included at the beginning of the prompt. Although unconventional, this observation suggests that polite phrasing can influence the model’s adherence to prompt instructions in certain scenarios.

\section{Conclusion and Future Work }
The overall integrated approach used by ourselves suggests that a proper fine tuning with augmented datasets provide comparatively a slight better result. The overall evaluation score would improve provided the model is fine tuned with Bangla Mathematics dataset. In our experiment. incorporating RAG which involves keyword search based similarity does not improve up to satisfactory level.Model performance may be better in case of using better retrieval models for RAG.Same goes for our large language model too. The more higher parameter a language model is, the more accurate result it would produce too. It is probable that a more curated and well organized dataset fine tuning may lead to more improvised and upgraded result in future.

\begin{table}[!htbp]
\centering
\scriptsize 
\renewcommand{\arraystretch}{1.8} 
\setlength{\tabcolsep}{10pt} 
\begin{tabular}{|p{4cm}|p{5cm}|p{5cm}|}
\hline
\textbf{Aspect} & \textbf{Observation} & \textbf{Recommendation} \\
\hline
Problem Categorization 
& Categorizing problems into types (e.g., Number Theory, Geometry) improves accuracy 
& Group problems by type for tailored prompts \\
\hline
Number Theory Prompts 
& Model performs better with brute force prompts for number theory 
& Include brute force hints in prompts for Number Theory \\
\hline
Geometry Prompts 
& Coordinate geometry approaches improved model performance in some cases 
& Prompt with coordinate geometry conversion hints for Geometry \\
\hline
Combinatorics and Functional Equations 
& Dynamic programming prompts enhance responses for combinatorial, functional, and recursive problems 
& Use dynamic programming prompts for these problem types \\
\hline
Multilingual Reasoning 
& Model performs better in English than in Bangla; repetitive reasoning improves results 
& Use both Bangla and translated English queries for repetitive reasoning \\
\hline
Politeness in Prompts 
& Adding “please” to the prompt sometimes results in stricter adherence 
& Consider polite phrasing like “please” in prompts to increase adherence \\
\hline
\end{tabular}
\vspace{10pt}
\caption{Observations and Recommendations for Enhancing AI Math Problem Solving}
\label{table:observations_recommendations}
\end{table}

\section{Dataset Availability}
\label{dataset_availability}

All the mentioned datasets can be downloaded from \href{https://docs.google.com/document/d/1Dvf6m7eRStmQh5kUONXF2DRMGfurXmxQptCiLuIz3Y8/edit?usp=sharing}{\textbf{here}}.

\section{Model Availability}
\label{model_availability}

All the fine-tuned model can be downloaded from \href{https://docs.google.com/document/d/1H-x0LZqL5vxq_Y8z9Pv5iePfvRxmz55_pnj720dW9js/edit?usp=sharing}{\textbf{here}}.



\vspace{12pt}

\end{document}